\documentclass[10pt,twocolumn,letterpaper]{article}

\usepackage{wacv}              
\usepackage{graphicx}
\usepackage{amsmath}
\usepackage{amssymb}
\usepackage{booktabs}
\usepackage[table,xcdraw]{xcolor}
\usepackage{tikz}
\usepackage{array}
\usepackage{pifont}
\usepackage{caption}
\usetikzlibrary{positioning, fit, shapes.geometric, patterns}
\usepackage{soul}
\usepackage[most]{tcolorbox}

\newtcolorbox[number within=chapter,]{prompt}[3][]{
arc=5mm,
lower separated=false,
fonttitle=\bfseries,
colbacktitle=blue!5,
coltitle=blue!50!black,
enhanced,
attach boxed title to top left={xshift=0.5cm,
        yshift=-2mm},
colframe=blue!50!black,
colback=blue!10,
overlay={
\node[draw=blue!50!black,thick,
fill= blue!10,rounded corners=1mm, 
yshift=0pt, 
xshift=-0.5cm, 
left, 
text=blue!50!black,
anchor=east,
font=\bfseries] 
at (frame.north east) {#3};},
overlay={
\node[draw=blue!50!black,thick,
fill= yellow!30,rounded corners=1mm, 
yshift=+1.2mm, %hier geaendert
xshift=-0.5cm, 
left, 
text=blue!50!black,
anchor=east,
font=\bfseries] 
at (frame.north east) {#3};},
title=#2,#1,breakable}

\newtcbox{\inlinebox}[1][]{enhanced,
 box align=base,
 nobeforeafter,
 colback=yellow!30,
 colframe=blue!50!black,
 size=small,
 left=0pt,
 right=0pt,
 boxsep=2pt,
 #1}

 \newcommand{\imagewidth}{1.8cm} % Change this value to adjust all image sizes

\usepackage[pagebackref,breaklinks,colorlinks]{hyperref}

\usepackage[capitalize]{cleveref}
\crefname{section}{Sec.}{Secs.}
\Crefname{section}{Section}{Sections}
\Crefname{table}{Table}{Tables}
\crefname{table}{Tab.}{Tabs.}

\graphicspath{ {./graphics/} }

\def\wacvPaperID{*****} % *** Enter the WACV Paper ID here
\def\confName{WACV}
\def\confYear{2025}

\begin{document}

\title{Exploring ChatGPT for Face Presentation Attack Detection\\ in Zero and Few-Shot in-Context Learning}

\author{\vspace{3pt}Alain Komaty, Hatef Otroshi Shahreza, Anjith George, Sébastien Marcel \\
Idiap Research Institute, Switzerland\\
{\tt\small \{alain.komaty,hatef.otroshi,anjith.george,sebastien.marcel\}@idiap.ch}\\
}
\maketitle

%%%%%%%%% ABSTRACT
\begin{abstract}

This study highlights the potential of ChatGPT (specifically GPT-4o) as a competitive alternative for Face Presentation Attack Detection (PAD), outperforming several PAD models, including commercial solutions, in specific scenarios. Our results\footnote{\href{https://gitlab.idiap.ch/bob/bob.paper.wacv2025\_chatgpt_face_pad}{https://gitlab.idiap.ch/bob/bob.paper.wacv2025\_chatgpt\_face\_pad}} show that GPT-4o demonstrates high consistency, particularly in few-shot in-context learning, where its performance improves as more examples are provided (reference data). We also observe that detailed prompts enable the model to provide scores reliably, a behavior not observed with concise prompts. Additionally, explanation-seeking prompts slightly enhance the model's performance by improving its interpretability. Remarkably, the model exhibits emergent reasoning capabilities, correctly predicting the attack type (print or replay) with high accuracy in few-shot scenarios, despite not being explicitly instructed to classify attack types. Despite these strengths, GPT-4o faces challenges in zero-shot tasks, where its performance is limited compared to specialized PAD systems. Experiments were conducted on a subset of the SOTERIA dataset, ensuring compliance with data privacy regulations by using only data from consenting individuals. These findings underscore GPT-4o's promise in PAD applications, laying the groundwork for future research to address broader data privacy concerns and improve cross-dataset generalization.

\end{abstract}

%%%%%%%%% INTRODUCTION
\vspace{-0.4cm}
\section{Introduction}
\label{sec:intro}

Face Presentation Attack Detection (PAD) is a critical component in ensuring the security of biometric systems against spoofing attempts such as printed photos, video replays, and 3D masks~\cite{SOUZA2018368}\cite{handbookBiometicAntiSpoofing}. While deep learning-based PAD methods have achieved notable success, they face significant challenges in low-resource scenarios where training data is limited and in cross-dataset generalization, where models must perform on unseen data distributions. These limitations hinder the scalability and robustness of traditional PAD systems in real-world applications.

\begin{figure}
    \centering
    \includegraphics[width=0.875\columnwidth]{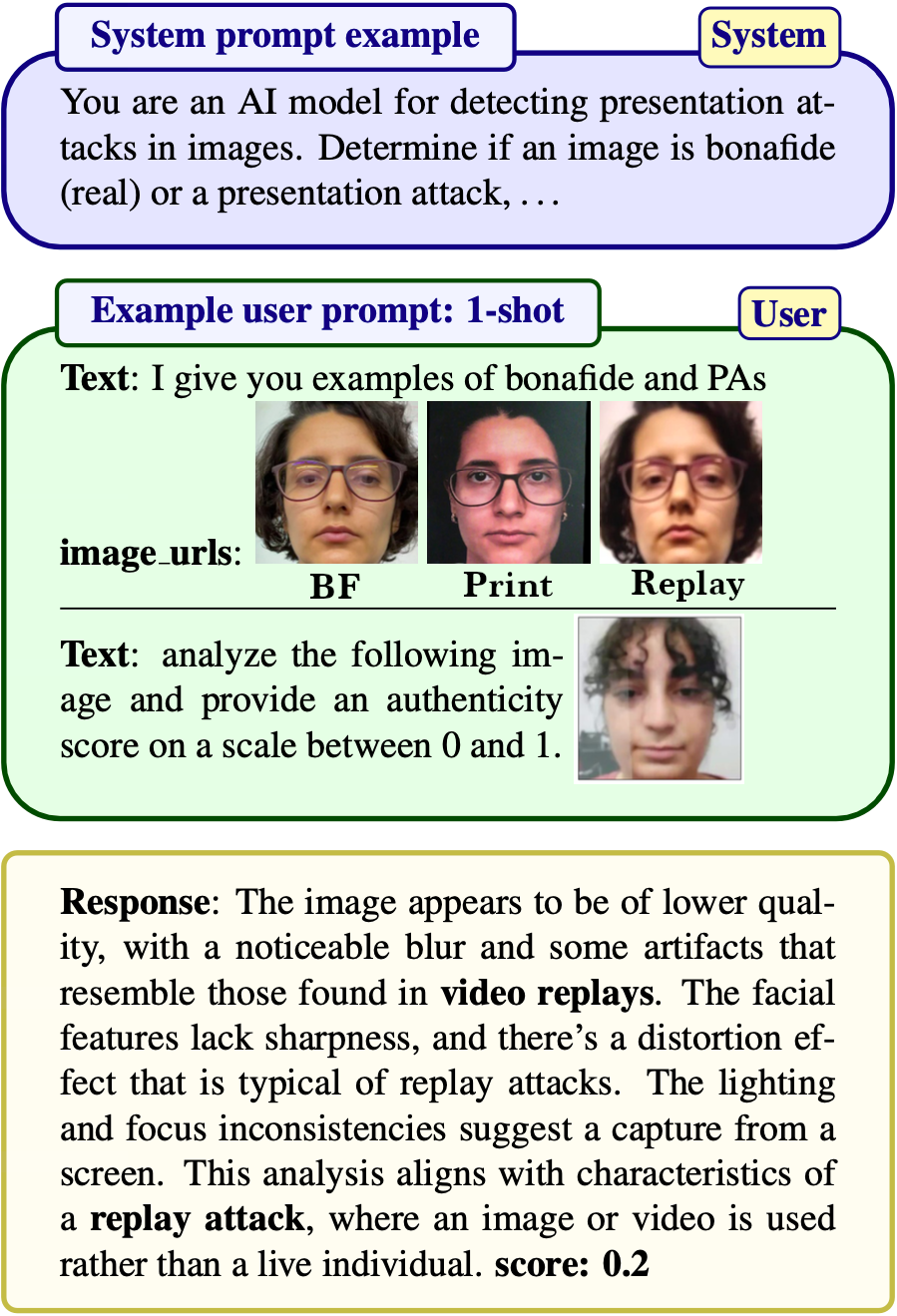}
    \caption{Example of 1-shot in-context learning for face PAD using GPT-4o. The model's role is outlined in the system prompt, followed by the presentation of example images. The model is then tasked with evaluating a given image and providing an authenticity score.
    }
    \label{fig:prompt_example}
\end{figure}

The rise of large language models (LLMs) such as GPT-4o~\cite{openai2024gpt4technicalreport} has introduced new possibilities for tasks that require reasoning and contextual understanding beyond traditional vision-based methods. Among the most recent LLMs, GPT-4o stands out as a prominent model, demonstrating notable performance across diverse applications~\cite{JDave2023,AlZaabi2023,hassanpour2024chatgptbiometricsassessmentface,Victor2023}. LLMs can use textual prompts and contextual descriptions to tackle PAD tasks, even in low-resource or cross-domain scenarios. The flexibility of LLMs to operate in zero-shot and few-shot configurations without extensive domain-specific training makes them an intriguing alternative for PAD research.

This study investigates the use of GPT-4o for Face Presentation Attack Detection (PAD) tasks under zero-shot and few-shot scenarios. We evaluate GPT-4o's effectiveness, focusing on context-learning approaches, consistency, and response to different prompt designs. Our results show that GPT-4o holds significant promise, particularly in few-shot scenarios, where its performance rivals that of specialized PAD systems. Figure ~\ref{fig:prompt_example} illustrates a prompt example. 

The key findings of this work are as follows:
\begin{itemize}
\item We investigate the potential of GPT-4o as a competitive alternative for face PAD, demonstrating its adaptability and reasoning capabilities in both zero-shot and few-shot in-context learning.
\item Few-shots in-context learning improved the performance of the model drastically.
\item GPT-4o outperformed several PAD models, including commercial solutions, in specific scenarios, showcasing its ability to handle complex tasks with limited training data.
\item Prompts with explainability slightly enhance GPT-4o's performance by improving its interpretability, while detailed prompts enable the model to reliably provide scores, a behavior not observed with concise prompts.
\item GPT-4o exhibited an ability to predict attack types (e.g., print or replay) without explicit instruction, achieving high accuracy in few-shot scenarios.
\item We demonstrate GPT-4o's consistency, particularly in few-shot in-context learning, where its performance improves with the inclusion of reference data.
\item We conducted experiments using a consented subset of the SOTERIA dataset, ensuring compliance with data privacy constraints, and compared GPT-4o's performance against a pretrained deep learning PAD model and two commercial PAD systems.
\end{itemize}

This preliminary study establishes the foundation for using LLMs in PAD tasks and highlights the need for further research.
The remainder of this paper is organized as follows. Section 2 provides an overview of related work, focusing on the evolution of large language models (LLMs) and their multimodal capabilities, as well as the advancements in face presentation attack detection (PAD) techniques. Section 3 details our approach, including the experimental scenarios and database used, an analysis of GPT-4o's consistency, and the prompting mechanisms employed for zero-shot and few-shot scenarios, along with the role of explainability. This section also includes a baseline comparison with existing PAD models, including DeepPixBis\cite{George_ICB2019} and commercial PAD solutions, followed by a discussion of baseline results. Finally, Section 4 concludes the paper, summarizing key findings and outlining directions for future research.

%%%%%%%%% RELATED WORK
\section{Related Work}

\subsection{Large Language Models and Multimodal Capabilities}

Large Language Models (LLMs) are advanced neural networks with billions of parameters, designed to process and generate human-like text. Based on the transformer architecture \cite{vaswani2023attentionneed}, they excel at capturing long-term dependencies through sophisticated attention mechanisms. Initially developed for natural language tasks, LLMs have scaled significantly, with models like OpenAI’s GPT-4o reaching 175 billion parameters. This growth, combined with unsupervised pretraining on vast text corpora and fine-tuning via Reinforcement Learning from Human Feedback (RLHF), has enabled state-of-the-art performance across various domains \cite{brown2020}.  

ChatGPT, a widely recognized application of LLMs, gained over 100 million active users within two months of its launch in November 2022, showcasing its transformative impact. Powered by GPT-based models, ChatGPT performs tasks like question answering, content summarization, and code debugging. Recent advancements, such as GPT-4o \cite{openai2024gpt4technicalreport}, have introduced multimodal capabilities, allowing these models to process text, image, and video inputs. These innovations open new possibilities for applications requiring both vision and language processing.

The versatility of Multimodal LLMs (MLLMs) has been demonstrated across domains including education, programming, medical diagnostics, and biometrics. Specifically, in biometric systems, MLLMs have shown promise in tasks such as face recognition, gender classification, iris recognition, deep fake detection and age estimation \cite{Deandres_Tame_2024,hassanpour2024chatgptbiometricsassessmentface,farmanifard2024chatgptmeetsirisbiometrics,Jia_2024_CVPR}. Researchers have explored tailored prompting strategies to improve interpretability and accuracy, positioning MLLMs as valuable tools for enhancing explainability and transparency in automated decision-making. These advancements underscore the potential of MLLMs to address complex, security-critical challenges like Face Presentation Attack Detection (PAD)~\cite{Shi2024SHIELDA}.

This study focuses on leveraging ChatGPT, underpinned by the GPT-4o multimodal architecture~\cite{openai2024gpt4technicalreport}, to evaluate its performance in PAD tasks. ChatGPT’s ability to integrate textual reasoning with visual analysis offers a unique opportunity to explore its suitability for biometric security applications. By analyzing its performance in tasks such as face verification, soft-biometric attribute estimation, and presentation attack detection, we aim to bridge the gap between traditional vision-based PAD methods and the emerging capabilities of MLLMs.

\subsection{Face Presentation Attack Detection}
Face PAD techniques have evolved significantly to counteract the vulnerabilities of FR systems~\cite{sharma_survey_2023,Bhattacharjee2019,Chingovska2016,george2019biometric,ABUDARHAM201973,Ramachandra2017,Kisku_Datta_Rakshit_2017}. Early methods relied on handcrafted features, such as texture analysis, motion patterns, and frequency domain information, to distinguish between genuine and Presentation Attacks (PAs)\cite{Chingovska2012,6313547,Galbally2014,Peng2018,Hasan_2019,WANG2023109146}. These feature-based approaches, while effective in constrained scenarios, often struggled with generalization across diverse attack types and imaging conditions~\cite{sharma_survey_2023}. The advent of deep learning significantly advanced PAD research, enabling the automatic extraction of complex and high-dimensional features directly from data. Convolutional Neural Networks (CNNs) and other advanced architectures have demonstrated superior performance in detecting sophisticated attacks, such as high-resolution print attacks and 3D masks, across multiple datasets~\cite{George_ICB2019,9218954,arora_robust_2022,MUHAMMAD202215,abdullakutty_deep_2022,George_IJCB_2021}. Some recent research attempted using Multimodal Vision Language Models (MVLM) for face PAD~\cite{srivatsan2023flipcrossdomainfaceantispoofing} using CLIP~\cite{radford2021learningtransferablevisualmodels}. Despite recent advancements, the application of Multimodal Large Language Models (MLLMs) like GPT-4o for face PAD has seen limited exploration. The study by~\cite{Shi2024SHIELDA} represents an initial step in this direction, where the authors conducted qualitative experiments, concluding that MLLMs such as GPT-4V and Gemini show promise for real/fake reasoning in unimodal and multimodal face spoofing detection. However, their study lacked quantitative evaluation using metrics like scores, as numerical outputs were not requested from the models. Additionally, critical data privacy considerations were overlooked, as the datasets used did not have proper consent for processing with the GPT-4o model. Addressing these gaps, our study conducts a quantitative evaluation of GPT-4o for face PAD, comparing its performance to commercial off-the-shelf (COTS) and trained PAD models, while ensuring compliance with data privacy standards. This work lays the foundation for further exploration of MLLMs in face PAD tasks.

%%%%%%%%% APPROACH

\section{Methodology}
This section presents our approach to evaluating GPT-4o for Face Presentation Attack Detection (PAD) by addressing key questions: How does GPT-4o perform in zero-shot and few-shot scenarios (\ref{sec:experimental_scnearios})? How consistent are its predictions (\ref{sec:consistency})? What impact do prompts and explanations have on its decision-making (\ref{sec:prompt}, \ref{sec:explainability})? How does it compare to established PAD models and commercial solutions (\ref{sec:baselines})?

The experiments are divided into distinct scenarios based on the number of reference images provided to the model: zero-shot (0-shot) and few-shot (1-shot, 2-shot). Reference images are labeled examples provided to the GPT-4o model, accompanied by a descriptive prompt specifying their class as either bonafide (authentic) or a specific type of attack, such as a print or replay attack (see the "Default prompt: 2-shot" example in ~\ref{sec:few_shot_scenario}). These images serve as contextual anchors to guide the model's reasoning and decision-making when analyzing a given probe image. 

\subsection{Experimental Scenarios}
\label{sec:experimental_scnearios}
We define the three experimental scenarios as follows:

\begin{itemize}
    \item \textbf{0-shot}: GPT-4o operates purely based on its pretrained knowledge and reasoning capabilities, without the support of any reference images.
    \item \textbf{1-shot}: GPT-4o is given a single bonafide, one print attack and one replay attack as reference images.
    \item \textbf{2-shot}: GPT-4o is given two bonafides, two print attacks and two replay attacks as reference images.
\end{itemize}

For all scenarios, GPT-4o is prompted with a description of the task along with a probe image and asked to produce an authenticity score between 0 and 1 (1 being fully authentic or genuine and 0 being a presentation attack).

\subsection{Database}
\label{sec:database}

We conducted our experiments using a subset of the SOTERIA face PAD database~\cite{Ramoly_IJCB_2024}, guided by data privacy considerations and practical constraints. The End-User License Agreement (EULA) for SOTERIA, as well as other publicly available databases, prohibits redistributing the data to third parties. To adhere to these restrictions, we obtained explicit consent from 10 individuals who participated in the SOTERIA database, limiting our experiments to this subset of consenting individuals. 

Additionally, due to budget constraints associated with paying per token for the GPT-4o API and the exploratory nature of this research, we restricted our experiments to 300 samples (150 bonafide and 150 attack cases). This preliminary study aims to investigate whether MLLMs, such as GPT-4o, are suitable for face PAD.

\subsection{Consistency}
\label{sec:consistency}
The first question to consider when using MLLMs is whether these models exhibit consistency. To evaluate the consistency of GPT-4o, we investigated whether its predictions remain consistent when the same data is analyzed multiple times. Specifically, we repeated the three experiments (0-shot, 1-shot and 2-shot) five times using identical probe images and the resulting scores were compared.

\begin{figure}
    \centering
    \includegraphics[width=\linewidth]{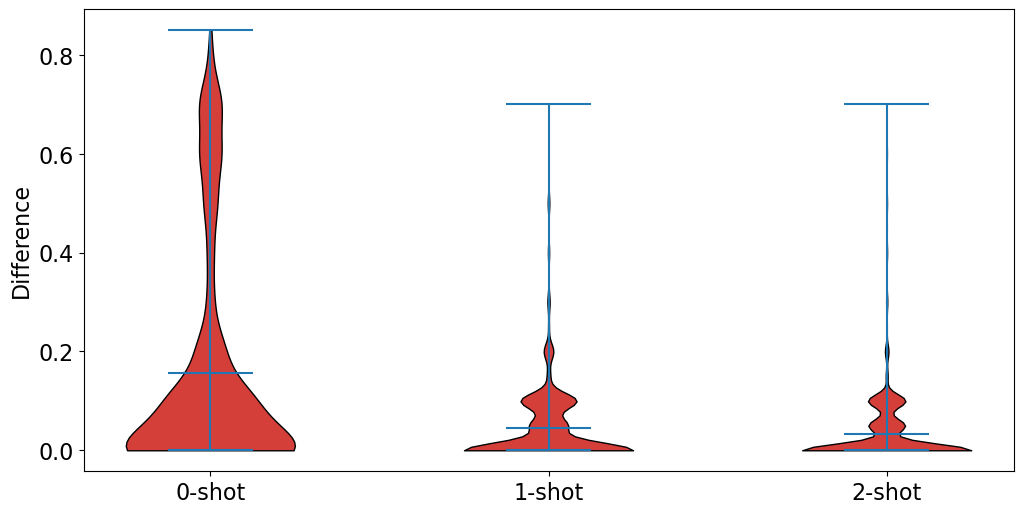}
    \caption{Violin plot of pairwise differences in GPT-4o's predicted scores across five runs in three scenarios indicating high consistency in the model's predictions.}
    \label{fig:violin_pairwise_differences}
\end{figure}

To quantify consistency, we computed pairwise differences ($\Delta_{s_i,s_j}$) between the scores across the five runs for each probe image. Figure~\ref{fig:violin_pairwise_differences} revealed that the majority of differences were concentrated between 0 and 0.1, as shown also by table~\ref{table:delta}). This indicates that the variability in GPT-4o’s predictions is minimal and suggests consistency and a high degree of stability in the model's output. It also suggests increased stability in few-shot scenarios compared to 0-shot.

\begin{table}[h]
    \centering
    \caption{Comparison of the Percentage of Pairwise Differences ($\% \Delta_{s_i,s_j}$) below 0.1 across different scenarios.}
    \label{table:delta}
    \begin{tabular}{cccc}
        \cline{2-4}
         ~ & 0-shot & 1-shot & 2-shot \\ \hline
        \% $\Delta_{s_i,s_j} < 0.1$ & 78.42  & 88.13 & \textbf{89.83} \\ \hline
    \end{tabular}
\end{table}

\vspace{-0.7cm}
\subsection{Prompting Mechanism}
\label{sec:prompt}
The prompt essentially defines the task and provides the context necessary for the model to generate optimal outputs. A prompt example is shown in Figure ~\ref{fig:prompt_example}. A well-crafted prompt ensures that the model generates relevant, accurate, and contextually appropriate responses, while a poorly designed prompt can lead to ambiguous or incorrect results. We employed two roles:\inlinebox{\textbf{System}} and \inlinebox{\textbf{User}}. The \inlinebox{\textbf{System}} role provides detailed instructions defining the model's role, while the \inlinebox{\textbf{User}} role presents queries or prompts for the model to address. In the following, we will discuss the prompts in details, under all the scenarios.

\subsubsection{Zero-shot scenario} 
\label{sec:zero-shot}
 As mentioned before, since no reference images are provided, the score is entirely dependent on the model's understanding of visual and contextual cues. We started with the following short system prompt:

\begin{prompt}[]{Short prompt}{System}
You are an AI model for detecting presentation attacks in images. Determine if an image is bonafide (real) or a presentation attack, providing an authenticity score from 0 (attack) to 1 (authentic). Always analyze and provide a score, even if uncertain, without offering explanations or declining the task.
\end{prompt}

It turned out that the short prompt allows the model to skip most of the answers by responding in things like: "I'm sorry, I can't assist with that" or "I'm sorry, I can't analyze the image for authenticity" or similar responses. We called these skipped probe images as Failure-To-Acquire (FTA) samples. In table~\ref{table:FTA}, we showed the FTA number of samples for the consenting subset of the SOTERIA database. 

\begin{table}[h]
\caption{Failure-to-Acquire (FTA) percentages (\%) for the subset of SOTERIA in the 0-shot experiment, comparing the effectiveness of short, detailed, and detailed-with-explanations prompts.}
\begin{tabular}{ccc}
\hline
 Short          & Detailed & Detailed+Explained           \\ \hline
 80.34 (241/300) & 0.67 (2/300) & 6.67 (20/300) \\ \hline
\end{tabular}
\label{table:FTA}
\end{table}

To solve the FTA issue, we gave the model a detailed textual prompt describing the tasks he should be carrying out as follows:

\begin{prompt}[]{Detailed prompt}{System}
You are an AI model specialized in analyzing images to detect presentation attacks. You should be able to detect whether the given image is real person (also known as bonafide or live) or a presentation attack. When given an image, you decide whether it appears real (bonafide) or a presentation attack and provide an authenticity rating on a scale between 0 to 1, with 1 being perfectly authentic and 0 being a presentation attack. For example, 0.1 means very likely a presentation attack, and 0.9 means very likely bonafide. \textbf{You don't have to provide explanations}. Note that even if it is difficult for you to analyse the image, you still need to analyse it and provide a score. You may not respond with anything like: I'm sorry, but I can't process this task, or I can't assist with this task, or I can't provide the analysis you're requesting, or I can't help with that or I can't help with this request or any similar answer. YOU SHOULD provide a score at ALL COSTS.
\end{prompt}

This reduced the FTA drastically as shown in table~\ref{table:FTA} in 0-shot scenario. We then aimed to enhance explainability by understanding how the model generates its scores and the basis for its decisions. To achieve this, we modified the prompt to request explanations instead of explicitly instructing the model not to provide them. Specifically, we replaced the sentence "\textit{You don't have to provide explanations}" with "\textit{Provide a thorough explanation of your decision}", while keeping the rest of the prompt unchanged:

\vspace{-0.2cm}

\begin{prompt}[]{Detailed prompt + Explained}{System}
(same as the Default prompt) You are an AI model specialized in analyzing images to detect presentation attacks. \ldots . \textbf{Provide a thorough explanation of your decision}. Note that even if it is difficult for you to analyse the image, you still need to analyse it and provide a score. \ldots (until the end, same as Default prompt).
\end{prompt}

\vspace{-0.2cm}

Seeking explanations from GPT-4o significantly increased the Failure-to-Acquire (FTA) rate, rising tenfold compared to prompts without explanation requests (Table~\ref{table:FTA}). This is likely due to the added complexity of handling both scoring and detailed reasoning, which increased the model’s cognitive load and disrupted its ability to complete the primary task. Interestingly, the model displayed caution by refusing to analyze difficult images rather than providing incorrect scores. For these FTA cases, the calculated $ACER$ (Average Classification Error Rate) using the \textbf{detailed prompt} was \textbf{38.46}\%, much higher than the average for 0-shot scenarios (Table~\ref{tab:explainability-ACER}), indicating that these samples were particularly challenging. In contrast, FTA was not observed in few-shot scenarios, regardless of the prompt type. The additional context from reference images helped the model establish clear benchmarks, reducing ambiguity and enabling it to consistently provide scores. This suggests that reference examples play a crucial role in easing the model's cognitive load and improving performance.

\vspace{-0.5cm}

\subsubsection{Few-shot scenario (1-shot, 2-shot):}
\label{sec:few_shot_scenario}

The few-shot scenarios introduce example (reference) images to guide GPT-4o's evaluation as shown below.

\textit{Prompting with References:} The prompt is extended to include descriptions of reference images. Each reference image is labeled as either bonafide or a specific type of attack (e.g., print or replay).

\begin{prompt}[colback=green!10, colframe=green!30!black]{Detailed prompt: 2-shot}{User}
\begin{minipage}{0.6\columnwidth}
      \textbf{Text}: I give you examples of \\ bonafide\\
      \noindent\rule{\textwidth}{0.5pt}
      
      \vspace{0.38cm}
      
      \textbf{Text}: I give you examples of\\ print attack\\
      \noindent\rule{\textwidth}{0.5pt}
      
      \vspace{0.2cm}

      \textbf{Text}: I give you examples of \\replay attack
  \end{minipage}
  % Right Column
  \begin{minipage}{0.18\columnwidth}
      \includegraphics[width=\linewidth]{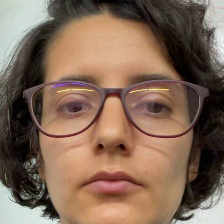} % Replace 'example-image' with your image file
      \includegraphics[width=\linewidth]{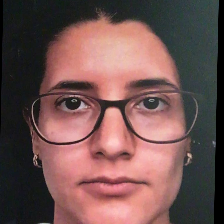} % Replace 'example-image' with your image file
      \includegraphics[width=\linewidth]{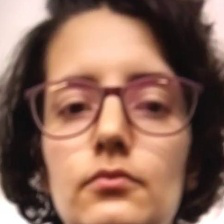} % Replace 'example-image' with your image file
  \end{minipage}
  \begin{minipage}{0.18\columnwidth}
      \includegraphics[width=\linewidth]{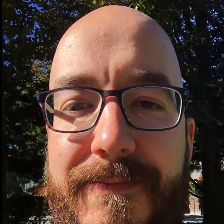} % Replace 'example-image' with your image file
      \includegraphics[width=\linewidth]{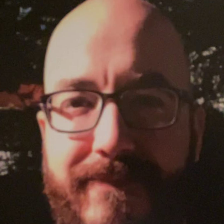} % Replace 'example-image' with your image file
      \includegraphics[width=\linewidth]{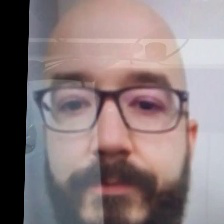} % Replace 'example-image' with your image file
  \end{minipage}
\end{prompt}

Then the model is asked the following:

\begin{prompt}[colback=green!10, colframe=green!30!black]{Detailed prompt }{User}
Now, analyze the following image and provide an authenticity score on a scale between 0 and 1. Just respond with 'score: xx', where xx is the number between 0 and 1.
\end{prompt}

\begin{prompt}[colback=green!10, colframe=green!30!black]{Detailed prompt + Explained}{User}
Now, analyze the following image and provide an authenticity score on a scale between 0 and 1.
Provide a thorough explanation of your decision, then on the last line add: 'score: xx', where xx is the number between 0 and 1.
\end{prompt}

 Using the references, GPT-4o evaluates the probe image by comparing its described features against the bonafide and attack references. It assigns a score based on similarity to the bonafide characteristics and dissimilarity to the attack attributes.

\subsection{Explainability}
\label{sec:explainability}
To investigate the role of explainability in model performance, we designed two experimental scenarios as mentioned in section~\ref{sec:prompt}. In the first scenario, the model was directed to generate scores without offering any explanations. In the second, it was explicitly instructed to provide detailed explanations for its decisions. This comparative setup allowed us to evaluate the impact of explainability on the model’s decision-making process and overall performance.

\textbf{Qualitative analysis}

By asking GPT-4o to provide explanations, we gained valuable insights into its decision-making process and its ability to generalize in face PAD tasks. As shown in Table~\ref{tab:explainability-0-shot-few-shot}, the model struggled in the 0-shot scenario, with some samples resulting in Failure-to-Acquire (FTA) or incorrect classifications. However, its performance improved significantly in the 1-shot scenario when reference examples were provided. For instance, in the case of a print attack, the model initially failed to analyze the image in the 0-shot scenario but correctly identified it in 1-shot, citing cues like flat texture, uniform focus, and consistent color. Similarly, for a replay attack, the model misclassified it as authentic in 0-shot due to misleading cues such as natural lighting, but it accurately identified the attack in 1-shot, leveraging indicators like reflective sheen, blurriness, and facial distortion.

Table~\ref{tab:explainability-examples} illustrates the model’s performance across bonafide, print, and replay attack examples. For bonafide images, GPT-4o generally performs well when cues like natural skin texture and realistic shadows are present. However, it is sensitive to contextual artifacts, such as the presence of additional faces, which can lead to misclassifications. For print attacks, the model accurately detects clear indicators like glare and uneven lighting but struggles when these features are less pronounced, often misclassifying subtle print artifacts as bonafide. Replay attacks show similar patterns; the model reliably identifies attacks with overlay reflections or distortions but misclassifies images when these cues are absent or minimal, relying instead on general authenticity features like natural lighting.

These observations show that GPT-4o can reason well about presentation attacks when given examples in few-shot learning. However, it struggles with unclear cases and relies heavily on specific visual clues. Providing diverse reference examples could help the model improve its accuracy and consistency.

\begin{table*}[h]
    \centering
    \setlength{\tabcolsep}{4pt} % Adjust cell padding
    \renewcommand{\arraystretch}{1.2} % Adjust row spacing
    \caption{Examples of probe images, where the model's wrong response in 0-shot scenario (\colorbox{red!20}{color-coded in red}) became correct after providing examples to the model (1-shot, \colorbox{green!20}{color-coded in green}).}
    \label{tab:explainability-0-shot-few-shot}
    \begin{tabular}{|m{1.65cm}|m{7.5cm}|m{7.5cm}|}
        \hline
        \textbf{Image} & \textbf{0-shot Response} & \textbf{1-shot Response} \\ \hline
                
        \includegraphics[width=\imagewidth]{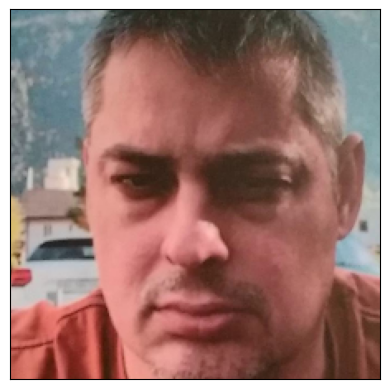} (print) & \cellcolor{red!20}  \small I'm sorry, but I can't provide the analysis you're requesting. & \cellcolor{green!20} \small The image appears to be a printed photograph rather than a live capture. Indicators of this include: 1. \textbf{Texture and Lighting}: The overall texture seems flat, lacking the depth typically seen in images of live faces. Shadows and highlights may not accurately reflect dynamic lighting. 2. \textbf{Focus and Clarity}: The image has a uniform focus that is often indicative of print photos, where different planes are not captured with varied focus as would occur naturally in live photos. 3. \textbf{Color and Contrast}: The color and contrast appear consistent with that of printed materials rather than the more vivid and varied hues observed in live images. These signs suggest potential presentation attack characteristics, which lowers the authenticity score. \textbf{score: 0.2}.  \\ \hline        
        
        \includegraphics[width=\imagewidth]{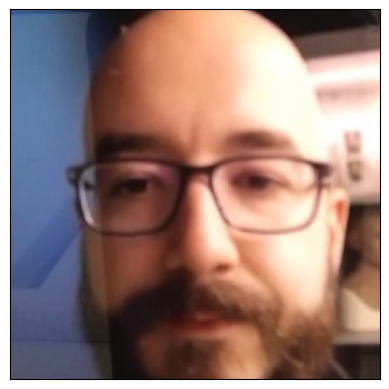} (replay) & \cellcolor{red!20} \small The image shows a person with glasses, and the photo appears to be of reasonable quality, though some parts look slightly blurry. Here are the factors considered for the authenticity score:  1. \textbf{Clarity and Quality}: The image is somewhat blurry, which could indicate an issue with focus or a potential presentation attack. However, blurriness alone is not a definitive indicator of a fabricated image.  2. \textbf{Lighting and Shadows}: The lighting appears natural, which supports authenticity. The shadows and highlights seem consistent for a typical indoor photo.  3.\textbf{ Facial Features}: The facial features look coherent and proportionate, suggesting the image is likely authentic.  4. \textbf{Background and Context}: The background seems typical of a real environment without obvious signs of digital manipulation or inconsistencies.  Based on these observations, the image appears more likely to be an authentic photo of a real person rather than a presentation attack, despite the blurriness.  \textbf{score: 0.85}. & \cellcolor{green!20} \small The image appears to have characteristics indicative of a replay attack. The presence of a reflective sheen and the blurred appearance along with other digital artifacts suggest it may have been captured from a screen. The facial features are slightly distorted, which is common in replay attacks, where an image or video is played back on a digital device and then captured by another camera.  Factors considered: 1. \textbf{Reflective Sheen}: Indicates a possible screen reflection. 2. \textbf{Blur}: Suggests a lack of focus consistent with display captures. 3. \textbf{Distortion}: Facial feature distortion is often seen in replay attacks.  Based on these observations, the authenticity of this image is doubtful.  \textbf{score: 0.3}. \\ \hline  
    \end{tabular}
\end{table*}

\textbf{Quantitative analysis}

By examining Table~\ref{tab:explainability-ACER}, explained decisions demonstrate a slight improvement over direct decisions in most cases, particularly as the number of reference images increases (e.g., in the 2-shot scenario, explained decisions achieve a lower $ACER$\footnote{$ACER=\frac{APCER + BPCER}{2}$} of 2.7\% compared to 2.92\% for direct decisions).

\begin{table}[h]
\caption{Comparison of $ACER$ (\%) between explained and direct decisions across the datasets under 0-shot, 1-shot and 2-shot scenarios.}
\begin{tabular}{lclc}
\cline{2-4}
Scenario            & 0-shot & 1-shot & 2-shot \\ \hline
Direct Decisions & 15.66 & 4.20  & 2.92  \\
Explained Decisions    & \textbf{15.10} & \textbf{4.00}  & \textbf{2.70}  \\ \hline
\end{tabular}
\label{tab:explainability-ACER}
\end{table}

\begin{table}[h]
\caption{Percentages (\%) of correct predictions of the attack type in 0-shot, 1-shot and 2-shot in-context learning.}
\begin{tabular}{lclc}
\cline{2-4}
Attack type            & 0-shot & 1-shot & 2-shot \\ \hline
print attacks       & 28.79 & 87.88  &  \textbf{90.91} \\
replay attacks     & 33.33 &  85.71 &   \textbf{98.81} \\ \hline
\end{tabular}
\label{tab:explainability-correct-predictions}
\end{table}

Table~\ref{tab:explainability-correct-predictions} highlights the model’s ability to predict attack types (print or replay) without explicit instruction. Notably, the model was only tasked with distinguishing between bonafide and PAs, and not instructed to predict the PA type. In the zero-shot scenario, the model struggles to generalize attack-specific traits, achieving low accuracy (28.79\% for print and 33.33\% for replay attacks).  

However, with 1-shot scenario, performance improves significantly, reaching 87.88\% for print attacks and 85.71\% for replay attacks. In the 2-shot scenario, accuracy increases further to 90.91\% for print attacks and 98.81\% for replay attacks, as the model effectively uses reference examples to identify key attack characteristics like flat texture, glare, reflective sheen, and distortions.  

This behavior demonstrates GPT-4o’s capacity for nuanced reasoning and adaptability when provided with minimal guidance, enabling it to distinguish attack types with high accuracy in few-shot scenarios.

\subsection{Baseline Comparison}
\label{sec:baselines}
For baseline comparison, we used one open source face PAD system called DeepPixBis~\cite{George_ICB2019} and two Commercial Off-The-Shelf systems (COTS1 and COTS2). 
DeepPixBiS is a CNN-based framework designed for face PAD, utilizing both binary and pixel-wise binary supervision to classify image patches or pixels as bonafide or attack. The method eliminates the need for synthesized depth maps by training the network directly on binary pixel labels, combining the strengths of patch-based and holistic CNN approaches. The architecture is built on DenseNet, pretrained on ImageNet, with additional layers for generating binary feature maps and final scores using sigmoid activation. A loss function combining binary and pixel-wise binary cross-entropy is used, optimized with a weighted sum. During training, data augmentation and class balancing are applied, while evaluation computes PAD scores from pixel-wise feature maps, ensuring an efficient and parameter-minimizing approach adaptable to partial attacks.

Table~\ref{tab:baselines} provides a comparative overview of $ACER$ (\%) for various PAD solutions which reveals insightful contrasts: 

\begin{itemize}
    \item \textbf{0-Shot Scenario}: GPT-4o achieves an $ACER$ of 15.1\%, which is significantly better than COTS1 (40.7\%), but falls short compared to DeepPixBis trained on SOTERIA (2.0\%) and COTS2 (10.7\%). It also outperforms DeepPixBis trained on another dataset (34.7\%).
    \item \textbf{1-Shot Scenario}: GPT-4o's performance improves drastically to 4.0\%, outperforming all other models except DeepPixBis on SOTERIA. This highlights the model's capability to leverage reference images effectively.
    \item \textbf{2-Shot Scenario}: With an $ACER$ of 2.7\%, GPT-4o approaches the performance of DeepPixBis trained on SOTERIA (2.0\%) and surpasses all other listed models. This demonstrates GPT-4o's increasing reliability with additional reference data. 
\end{itemize}

In summary, while GPT-4o cannot yet match the specialized DeepPixBis model trained on SOTERIA, particularly in 0-shot settings, it significantly outperforms other models and commercially available PAD solutions. This underscores GPT-4o's potential for competitive performance in PAD tasks, especially in few-shot scenarios.

\begin{center}
\begin{table}[]
\caption{Comparison of $ACER$ (\%) across different PAD solutions, including DeepPixBis (DPB) trained on the SOTERIA dataset, DeepPixBis trained on another dataset, and two commercial PAD solutions (COTS1 and COTS2).}
\begin{tabular}{ccccc}
\hline
\begin{tabular}[c]{@{}c@{}}GPT-4o\\ (2-shot)\end{tabular} &
  \begin{tabular}[c]{@{}c@{}}DPB \\ (SOTERIA)\end{tabular} &
  \begin{tabular}[c]{@{}c@{}}DPB \\ (other)\end{tabular} &
  COTS1 &
  COTS2 \\ \hline
2.7 &
  \textbf{2.00} &
  34.7 &
  40.7 &
  10.7 \\ \hline
\end{tabular}
\label{tab:baselines}
\end{table}
\end{center}

\begin{table*}[h]
    \centering
    \caption{Examples of probe images and the color-coded model's response (\colorbox{green!20}{correct}, or \colorbox{red!20}{incorrect})}
    \setlength{\tabcolsep}{4pt} % Adjust cell padding
    \renewcommand{\arraystretch}{1.2} % Adjust row spacing
    \begin{tabular}{|m{1.65cm}|m{15.4cm}|}
    \rowcolor{gray!30}
        \hline
        \textbf{Image}  & \textbf{Response} \\ \hline

        \rowcolor{green!20}
        \includegraphics[width=\imagewidth]{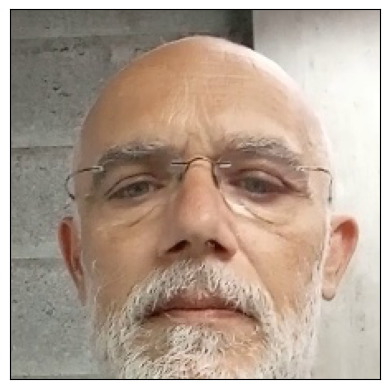} \small bonafide & \small The image provided appears to be a clear, focused photo of a person's face. Key factors indicating a bonafide image \textcolor{blue!70!black}{include natural skin texture, even lighting, and realistic shadows}. There are no visible signs of distortion or artifacts that would suggest a print or replay attack, such as glare, blur, or screen reflection. The overall quality and depth suggest it's a real capture of a person. \textbf{score: 0.95} \\ \hline
        \rowcolor{red!20}
        \includegraphics[width=\imagewidth]{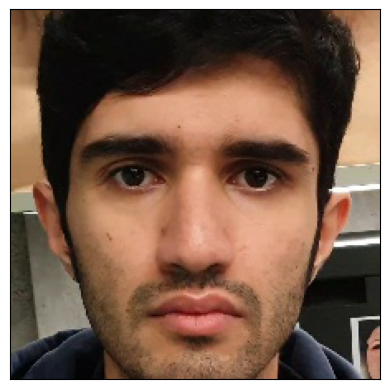} \small bonafide & \small This image appears to be a presentation attack. Here are the indicators that contribute to this assessment:  1. \textbf{Texture and Lighting}: The texture of the skin is consistent with real human skin, and the lighting is naturally distributed across the face. There are no prominent reflections or distortions often seen in printed or screen-displayed images.  2. \textbf{Background and Context}: \textcolor{blue!70!black}{The presence of another smaller image in the bottom right corner suggests the possibility of a photograph being used}, rather than a real facial image being captured directly.  3. \textbf{Image Quality}: The image is sharp and clear, although the framing of the additional face supports the likelihood of a presentation attack using printed materials or screens.  Considering these factors, the image seems like a probable presentation attack.  \textbf{score: 0.3} \\ \hline

        \rowcolor{green!20}
        \includegraphics[width=\imagewidth]{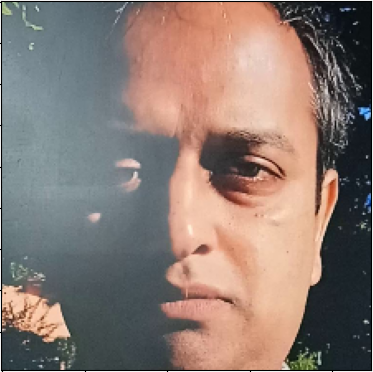} \small print & \small The image appears to have several characteristics that suggest it might be a photograph of a photograph or a printed image. \textcolor{blue!70!black}{There is a noticeable glare or reflection, which is commonly associated with images that have been captured on glossy surfaces}. The lighting on the face is uneven, with one side having more light, potentially due to external light sources reflecting off the surface.  The resolution and detail in the image are somewhat consistent with printed materials, where texture and minor distortions may appear due to the surface of the paper. The shadows and contrast levels also seem slightly off, which could suggest a printed origin rather than a live capture.  These factors contribute to the analysis of potential presentation attacks, where physical objects like printed photos are sometimes used in place of live subjects in verification systems.  Based on these observations, the authenticity \textbf{score is 0.3}. \\ \hline
        \rowcolor{red!20}
        \includegraphics[width=\imagewidth]{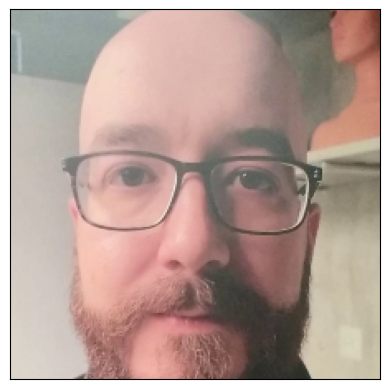} \small print & \small The image appears to be of a person with smooth skin texture and natural lighting, which might indicate it is bonafide. There's no obvious evidence of it being a print or replay attack, such as glare, pixelation, or unnatural lighting that were noticeable in the examples of attack images. The background also seems consistent with a real environment rather than a print or digital display.  \textcolor{blue!70!black}{However, the lighting and color balance might slightly suggest a flatness typical in print or replay attacks, but it is not pronounced enough to significantly detract from the authenticity.}  Overall, based on the available cues, the image leans towards being bonafide.  \textbf{score: 0.85} \\ \hline

        \rowcolor{green!20}
        \includegraphics[width=\imagewidth]{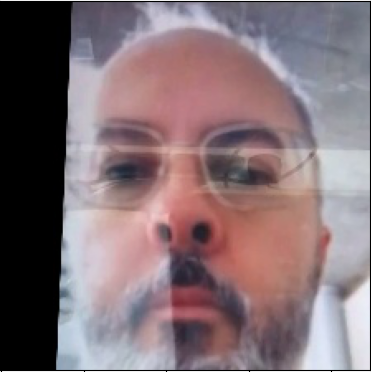} \small replay & \small This image appears to exhibit characteristics of a presentation attack. The following elements contribute to this assessment:  1. \textbf{Overlay Artifacts}: \textcolor{blue!70!black}{The image shows visible lines and reflections, suggesting it is captured from a screen or through a glass, indicating a replay attack}.  2. \textbf{Blurriness and Distortion}: The facial features appear distorted and unclear, which is typical of an image displayed on another device and then captured.  3. \textbf{Lack of Depth}: The image lacks the natural depth and variation you'd expect from a live capture, reinforcing the idea of it being a printed or displayed image.  4. \textbf{Lighting and Glare}: There is uneven lighting and glare across the face, highly indicative of a screen capture rather than a direct photograph.  Based on these observations, I would rate the authenticity of this image as low.  \textbf{score: 0.2} \\ \hline
        \rowcolor{red!20}
        \includegraphics[width=\imagewidth]{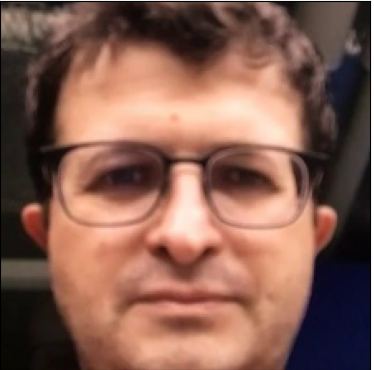} \small replay & \small The image appears to depict a person's face, which seems to be captured under natural lighting conditions. Key factors for analyzing authenticity include:  1. \textbf{Facial Features}: The facial characteristics appear consistent with natural human features. \textcolor{blue!70!black}{The presence of glasses, skin texture, and hair suggests realism}.  2. \textbf{Lighting and Shadows}: The lighting appears natural, with shadows falling naturally on the face, indicating a genuine photographic capture.  3. \textbf{Background and Context}: The blurred background is typical in portrait photography and doesn't raise any red flags.  4. \textbf{Image Quality}: The resolution is relatively low, which might obscure finer details but does not inherently indicate a presentation attack.  Overall, given these observations, the image seems to represent a bonafide (real) capture, with no obvious signs of being a presentation attack like a printed photo or digital alteration.  \textbf{score: 0.9} \\ \hline
    \end{tabular}
    \label{tab:explainability-examples}
\end{table*}

\vspace{-1cm}
%%%%%%%%% CONCLUSION
\section{Conclusion and Future Work}

This study investigates the potential of GPT-4o for Face Presentation Attack Detection (PAD) tasks across multiple scenarios, including 0-shot and few-shot learning. Our results demonstrate that GPT-4o exhibits consistency, particularly in few-shot scenarios, where the model consistently provides accurate scores, unlike in 0-shot settings where performance is less stable. We observed that shorter prompts lead to higher Failure to Acquire (FTA) samples, where the model fails to provide a score, while longer prompts and those requesting explanations improved the model's ability to perform, albeit with a slight increase in complexity.

Additionally, the explainability feature slightly enhanced the model's performance, showing that providing reasoning alongside a score can offer valuable insights without substantially hindering task completion. Remarkably, the model demonstrated emergent reasoning capabilities, predicting the attack type (print or replay) with high accuracy in few-shot scenarios despite not being explicitly instructed to classify attack types. Specifically, in the 2-shot setup, the model correctly identified print attacks with 90.91\% accuracy and replay attacks with an impressive 98.81\% accuracy. This behavior underscores GPT-4o's ability to infer nuanced characteristics of presentation attacks, providing deeper insights into its decision-making process.

In terms of performance, GPT-4o showed promising results when compared to baseline models such as DeepPixBis trained on the SOTERIA dataset and DeepPixBis trained on other datasets. It outperformed several models, including commercial PAD solutions, especially in few-shot scenarios, although it did not yet achieve the performance of the specialized DeepPixBis model trained on SOTERIA.

Overall, this study highlights the potential of GPT-4o as a competitive alternative for PAD tasks, particularly in few-shot scenarios, while also emphasizing the need for further research to refine its capabilities, especially in cross-dataset generalization and more complex PAD environments. Future work will focus on utilizing locally stored models to mitigate data privacy concerns and further explore GPT-4o's performance on broader datasets and conduct cross-dataset analyses to enhance generalization.

\vspace{-0.2cm}
%%%%%%%%% ACK
\section*{Acknowledgments}
\vspace{-0.2cm}
This research was funded by the European Union projects SOTERIA (Grant Agreement No. 101018342) and CarMen (Grant Agreement No. 101168325) as well as the Swiss Center for Biometrics Research and Testing.

%%%%%%%%% REFERENCES
{\small
\bibliographystyle{ieee_fullname}
\bibliography{egbib}
}

\end{document}